\pdfoutput=1

\documentclass[11pt]{article}

\usepackage[]{ACL2023}

\usepackage{times}
\usepackage{latexsym}
\usepackage{stfloats}
\usepackage[T1]{fontenc}

\usepackage[utf8]{inputenc}

\usepackage{microtype}

\usepackage{inconsolata}

\usepackage{enumitem} 
\usepackage{color}
\usepackage{soul}
\usepackage{booktabs}
\usepackage{amsmath,amssymb}
\usepackage{subcaption}
\usepackage{multirow}
\usepackage{graphicx}
\usepackage{float}

\title{Scenarios and Approaches for Situated Natural Language Explanations}

\author{Pengshuo Qiu$^{1}$, Frank Rudzicz$^{2,3}$, Zining Zhu$^{2,4}$\\
$^1$ Tohoku University, $^2$ Vector Institute for Artificial Intelligence, \\
$^3$ Dalhousie University, $^4$ Stevens Institute of Technology \\
  \texttt{qiu.pengshuo.t5@dc.tohoku.ac.jp}, \texttt{frank@spoclab.com}, \texttt{zzhu41@stevens.edu}}

\begin{document}
\maketitle
\begin{abstract}

\begin{table*}
\centering
\resizebox{\textwidth}{!}{%
\begin{tabular}{clll}
\hline
\multicolumn{1}{l}{\textbf{Explanandum}} &
  \textbf{Audiences} &
  \textbf{Desired Features} &
  \textbf{Explanations} \\ \hline
\multirow{3}{*}{\begin{tabular}[c]{@{}c@{}}Educational  
 \\ technology \\  can be meaningful\end{tabular}} &
  Students &
  \begin{tabular}[c]{@{}l@{}}Interested in engaging learning tools \\tailored to individual preferences\end{tabular} &
  \begin{tabular}[c]{@{}l@{}}Educational technology can provide \\a personalized and enjoyable learning experience \\through interactive resources...\end{tabular} \\ \cline{2-4} 
 &
  \begin{tabular}[c]{@{}l@{}}Teachers\end{tabular} &
  \begin{tabular}[c]{@{}l@{}}Interested in improving efficiency,\\ streamlining teaching tasks \end{tabular} &
  \begin{tabular}[c]{@{}l@{}}Educational technology can empower \\teachers to automate tasks...\end{tabular} \\ \cline{2-4} 
 &
  \begin{tabular}[c]{@{}l@{}}Parents\end{tabular} &
  \begin{tabular}[c]{@{}l@{}}Interested in seeking for\\ engaging resources \\helping their children with studying \end{tabular} &
  \begin{tabular}[c]{@{}l@{}}Educational technology can provide \\resources that parents can use to \\help their children study...\end{tabular} \\ \hline
\end{tabular}}
\caption{An example of a scenario in \textsc{SBE}.}
\label{citation-guide}
\end{table*}
Large language models (LLMs) can be used to generate natural language explanations (NLE) that are adapted to different users’ situations. However, there is yet to be a quantitative evaluation of the extent of such adaptation. To bridge this gap, we collect a benchmarking dataset, \textsc{Situation-Based Explanation}. This dataset contains 100 explanandums. Each explanandum is paired with explanations targeted at three distinct audience types-such as educators, students, and professionals-enabling us to assess how well the explanations meet the specific informational needs and contexts of these diverse groups e.g. students, teachers, and parents. For each ``explanandum paired with an audience'' situation, we include a human-written explanation. These allow us to compute scores that quantify how the LLMs adapt the explanations to the situations. On an array of pretrained language models with varying sizes, we examine three categories of prompting methods: rule-based prompting, meta-prompting, and in-context learning prompting. We find that 1) language models can generate prompts that result in explanations more precisely aligned with the target situations, 2) explicitly modeling an "assistant" persona by prompting "{\em You are a helpful assistant...}" is not a necessary prompt technique for situated NLE tasks, and 3) the in-context learning prompts only can help LLMs learn the demonstration template but can't improve their inference performance. \textsc{SBE} and our analysis facilitate future research towards generating situated natural language explanations.
\end{abstract}

\section{Introduction}
\begin{figure}
\centering
\includegraphics[scale=0.45]{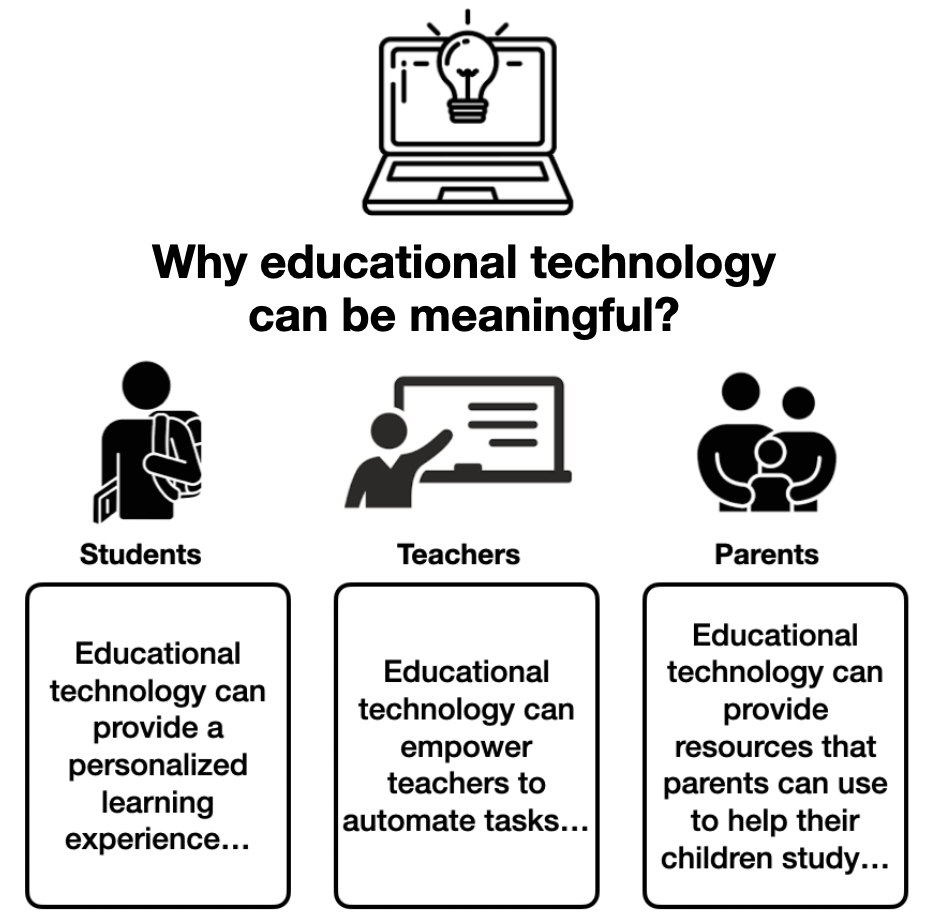}
\caption{Different audiences need different explanations.}
\end{figure}
Recently, LLMs have shown promising abilities to reason about complex phenomena and explain them in fluent natural languages. The produced natural language explanations (NLEs)\footnote{NLE is also termed ``free-text rationales'' in literature: \citet{deyoung-etal-2020-eraser,joshi-etal-2023-machine,chen-etal-2023-rev}, \textit{inter alla}. We consider the two terms synonymous.} can be highly accurate \citep{narang2020wt5}, informative \citep{wiegreffe-etal-2022-reframing}, plausible \citep{chan2022frame,marasovic-etal-2022-shot} and reasonably faithful \citep{lyu-etal-2023-faithful}. These desirable properties lead to wide potential applications that use LLMs as building blocks for explainer tools.

The explainer tools are closely relevant to the users, which make some properties particularly desirable, for example, helpful for answering unseen instances \citep{joshi-etal-2023-machine} and for fact-checking \citep{si2024large}. In this paper, we are particularly interested in situatedness: the explanations of the same phenomena can and should be tailored to the audience. This principle is well-established in the literature of psychology, education, and communication (e.g., the Cognitive Load Theory \citep{vanmerrienboer2005Cognitive}), and is prevalent in writing guides \citep{purdueTone,stephen2022Guide,cutts2020oxford}, even in government's writing guidelines \citep{administrationWrite}. 

Recent work by \citeposs{zhu2023Situated} explored using pretrained language models for generating situated NLEs, which are explanations adapted to the situations of different users. However, their work only involves rule-based adaptation methods, and lacks a quantitative evaluation of the extent of such adaptation. 

In this paper, we aim to bridge both gaps. We introduce a novel benchmarking dataset called \textsc{Situation-Based Explanation} (\textsc{SBE} for short). This dataset contains 100 explananda (concepts or phenomena to be explained), each paired with three potential audiences. For each unique combination of explanandum and audience, we provide a human-written explanation, allowing us to compute similarity scores and a matching score that quantify how well the language models adapt the explanations to the target situations.
Using \textsc{SBE}, we systematically evaluate the performance of various pretrained language models across three categories of prompting methods: rule-based prompting, meta-prompting, and in-context learning prompting. Through this analysis, we uncover the strengths and limitations of different prompting techniques in generating situated NLEs.
Our key contributions are:
\begin{enumerate}
\item We provide \textsc{SBE}, a dataset facilitating systematic study of the situated adaptation effects of NLE.
\item We quantify the effects of several prompting techniques for generating situated explanations with LLMs.

\end{enumerate}

By introducing \textsc{SBE} and rigorously evaluating the performance of various LLMs and prompting methods, this work paves the way for future research towards more effective and situationally appropriate natural language explanations.

\section{Related Work}

\paragraph{Natural language explanation}
Natural language explanations (NLEs) have been widely studied for various high-level reasoning tasks, such as inference \citep{camburu-etal-2020-make}, commonsense multiple-choice questions \citep{rajani-etal-2019-explain}, question-answering \citep{aggarwal-etal-2021-explanations}, and product recommendations \citep{10.1145/3340531.3411992}. \citet{NEURIPSDatasetsBenchmarks2021} provided a comprehensive review of datasets for NLEs, highlighting desirable properties of NLEs including simplicity \citep{LOMBROZO2007232}, clarity, and informativeness \citep{clinciu-etal-2021-study}. We consider a different property: the extent of adaptation to the situational contexts of the explanations' readers.

\paragraph{Human-centered explanation}
Despite the numerous methods proposed to make AI more explainable, most focus on mechanistic approaches, whose utilities are subject to increasing concerns that these methods fail to consider the cognitive and contextual needs of users. Multiple researchers have called for greater consideration of human factors when explaining AI \citep{Liao_Zhang_Luss_Doshi-Velez_Dhurandhar_2022, boyd-graber-etal-2022-human, yeung2020sequential, MILLER20191, ehsan2024Who, goyal2023What}. Building upon this line of research, we develop NLE methods that are technically accurate and effectively communicate with and meet the needs of human users.
Explainable recommendation is also a related research direction, where the personification of explanation would be beneficial for recommendation systems \citep{geng2022Improving,geng2022recommendation}. We consider a wide range of scenarios, and our findings can be applied to explainable recommendations.

\paragraph{Cultural and societal knowledge} 
This paper is related to the works about the pragmatics of LM-based communication tools. \citet{yerukola2024pope} considered the abilities to infer the speakers' intents. \citet{rao2024normad} considered the cultural adaptabilities of LLMs. \citet{liu2024Evaluating} found that LLMs are less steerable when personified to stances associated with apparently incongruous traits. We focus on daily-life scenarios without the cross-cultural differences.
Our work is also closely related to researches that assess the societal intelligence of LLM-based language generation systems, including \citet{wang2024sotopia} which simulated societal interactions between LLM-based agents. We focus on the situated adaptation of NLE.

\section{Data}
\begin{figure}
\centering
\includegraphics[scale=0.4]{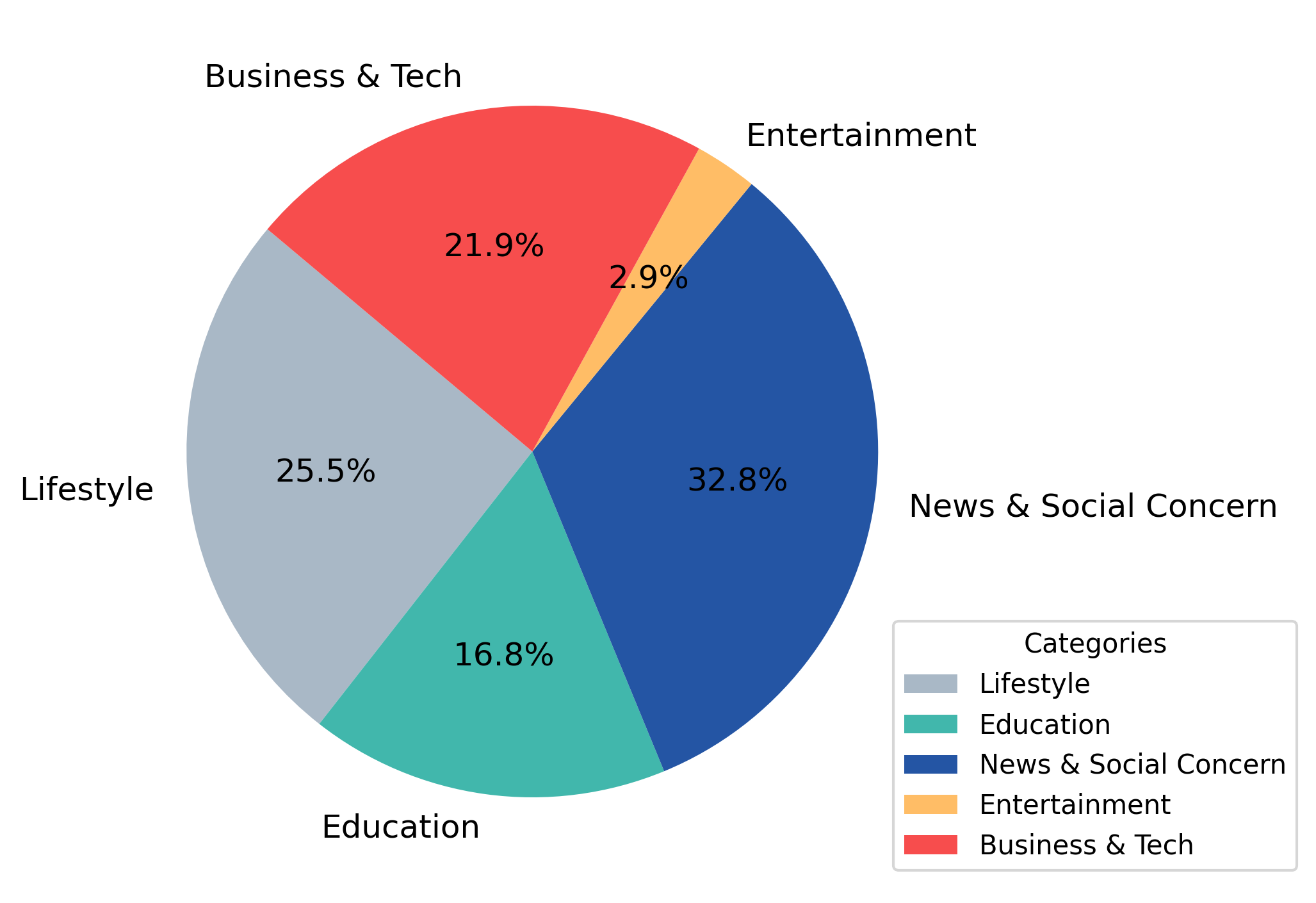}
\caption{The distribution of categories in \textsc{SBE}.}
\label{distribution}
\end{figure}

We identify 100 real-world scenarios across a broad spectrum of topics. Figure \ref{distribution} illustrates the distribution of categories in \textsc{SBE}. The categories include:
\begin{itemize}
    \item \textbf{Lifestyle}: Arts \& culture, diaries \& daily life, fashion \& style, fitness \& health, food \& dining, other hobbies, sports, travel \& adventure, youth \& student life.
    \item \textbf{Business \& Tech}: Business \& entrepreneurs, science \& technology.
    \item \textbf{Entertainment}: Celebrity \& pop culture, film, TV \& video, gaming, music.
    \item \textbf{Education}: Learning \& educational.
    \item \textbf{News \& Social Concern}: News \& social concern.
\end{itemize}
For each scenario, we pinpoint a central concept to be explained (explanandum) and crafted three distinct situations, each representing a potential audience with unique perspectives and concerns (desired feature). These audiences include foodies interested in exploring new and unique dishes, students navigating mental health challenges, and social media influencers looking to enhance their content engagement strategies.

To ensure the validity of our benchmarks, we systematically develop a set of explanations for each scenario. To reflect the diverse needs and backgrounds of the specified audiences, explanations are manually written by our research team. We ensure that each explanation not only addresses the central concept but also resonates with the specific interests and concerns of the audience. This approach provides us with a rich dataset of explanations that are both contextually relevant and varied across different domains. Utilizing these carefully crafted explanations as a benchmark, we evaluate the performance of LLMs in generating situated NLEs.

\section{Methods}
\subsection{Rule-based prompting methods}

\paragraph{Base prompt} 
\begin{center}
\fcolorbox{black}{gray!10}{\parbox{.9\linewidth}{
\textbf{Base}: \{explanandum\} because
}}
\end{center}

\paragraph{Specify the audience or the desired feature}
Whereas previous work often limited prompting modifications to either audience or context-specific features, our approach combines both. This dual focus enables a more precise tailoring of explanations to the audience's needs and the contextual nuances of the explanandum:

\begin{center}
\fcolorbox{black}{gray!10}{\parbox{.9\linewidth}{
\textbf{Audience Specification} (1A): Following is an explanation towards \{audience\}. \{explanandum\} because\\
\textbf{Desired Feature Specification} (1D): Following is an explanation about \{desired feature\}. \{explanandum\} because\\
\textbf{Audience and Feature Specification} (1AD): Following is an explanation towards \{audience\}, about \{desired feature\}. \{explanandum\} because
}}
\end{center}

This innovation, denoted as 1AD, represents an extension beyond traditional single-focus prompts, aiming to enhance the relevance and clarity of the generated explanations.

\paragraph{Adopt a persona}
\begin{center}
\fcolorbox{black}{gray!10}{\parbox{.9\linewidth}{
\textbf{Do not simulate} (2F): Following is an explanation towards \{audience\}. \{explanandum\} because\\
\textbf{Simulate the model as a helpful assistant} (2T): You are a helpful assistant explaining to \{audience\}. \{explanandum\} because\\
}}
\end{center}

\paragraph{Elicit the NLE with complete sentences}
\begin{center}
\fcolorbox{black}{gray!10}{\parbox{.9\linewidth}{
\textbf{Just use “because”} (3F): Following is an explanation towards \{audience\}. \{explanandum\} because\\
\textbf{Use a complete sentence} (3T): Following is an explanation towards \{audience\}: \{explanandum\}.
}}
\end{center}

By combining these three techniques, we construct a set of 12 distinct prompts (from 1A2F3F to 1D2T3T to 1AD2T3T).

To further enhance the adaptability and effectiveness of our prompting techniques, we introduce a set of new templates that have been empirically tested to optimize the models' performance. These templates are designed to integrate seamlessly with the refined prompting strategies, ensuring that each prompt not only cues the model for content generation but also aligns closely with requirements of the situated NLE task.

\subsection{Meta prompt}
To harness the capabilities of LLMs in generating contextually appropriate prompts, we employ a structured approach. This involves framing the prompt to explicitly include the intended audience and the specific features of interest, directing the model to tailor its response accordingly. Here is the format we propose for such prompts:

\begin{center}
\fcolorbox{black}{gray!10}{\parbox{.9\linewidth}{
You are a helpful assistant helping me write a prompt. I want to write a prompt to generate an explanation about why \{explanandum\} to \{audience\}, about \{desired feature\}. Give me the prompt directly.
}}
\end{center}

In our experiment, we tested prompts generated by GPT-3.5 on other models. Additionally, we evaluated each model's ability to generate its own prompts and then respond to them. This dual approach allowed us to not only assess the transferability of prompts across different models but also evaluate each model's capacity for self-driven contextual understanding and prompt formulation.

\subsection{In-context Learning Prompt}
In-context learning prompts serve as a powerful tool to enhance a model's ability to generate context-specific explanations. The suggested format for in-context learning prompts for audience1 in one situation is:

\begin{center}
\fcolorbox{black}{gray!10}{\parbox{.9\linewidth}{
Q: Following is an explanation towards \{audience2\}, about \{desired feature2\}. \{explanandum\} because\\
A: \{explanation2\}\\
Q: Following is an explanation towards \{audience3\}, about \{desired \ feature3\}. \{explanandum\} because\\
A: \{explanation3\}\\
Q: Following is an explanation towards \{audience1\}, about \{desired feature1\}. \{explanandum\} because\\
A:
}}
\end{center}

The rationale for choosing the "1AD2F3F" template over others for the in-context learning prompt was guided by specific performance metrics observed in our initial testing phase. Specifically, the "1AD2F3F" template showed superior performance in terms of similarity scores and matching score when compared to other templates. The decision to focus on a single template, rather than attempting to replicate all possible templates in our demonstration setting, was based on practical constraints and the desire to optimize the demonstration’s relevance and efficiency. 

Overall, we collect 16 prompts and evaluate  performance of several LLMs on a situated NLE task by each prompt. 

\section{Experiment setup}
\subsection{LLM explainers}
We use 5 LLMs, including GPT-3.5-turbo \cite{NEURIPS2022_b1efde53}, Pythia-2.8B \cite{10.5555/3618408.3618510}, LLaMa2-7B, LLaMa2-13B-chat \cite{touvron2023LLaMa}, and Yi-34B \cite{ai2024yi}, to generate situated NLEs.

\subsection{Evaluation}
In our experiments, we focus on two key metrics: similarity score and matching score. The similarity score measures the semantic similarity between the generated explanation and the desired explanation for a given situation. Matching score evaluates the overall suitability of the generated explanation to the given situation.

\paragraph{Similarity score}
To measure the similarity between two sentences, we employ a sentence embedding model, Sentence-BERT (SBERT) \citep{reimers-gurevych-2019-sentence}. Specifically, we use the all-MiniLM-L6-v2 model from the sentence-transformers library \citep{reimers-2019-sentence-bert}. The similarity between the two sentences is then computed as the cosine similarity between their embeddings. The score ranges from $-1$ to $1$, with higher values indicating greater semantic similarity between sentences. 

A higher similarity score indicates an LLM-generated explanation is more approximate to the human-annotated explanation in \textsc{SBE}. However, how ought one evaluate whether an LLM-generated explanation is suitable for the audience in the situation?

\paragraph{Matching Score}
Within our evaluative framework, we adopt a scoring methodology predicated on the cross-entropy loss function to quantify the congruence between explanations generated by the LLM and the target explanations. The formula \(\sum_{c=1}^{N} y_c \log(p_c)\) delineates the computation of loss for a multi-class classification task, where \(N\) signifies the number of classes. Herein, the loss aggregates the weighted negative logarithms of the predicted probabilities \(p_c\) across classes \(c\), with the weighting provided by the actual class indicators \(y_c\).

For the purpose of our experiment, we designate \(N=3\) to align with the triad of situational contexts within a singular scenario. Let \(c\) denote the situational index, and \(j\) represent the index for LLM-generated explanations. As we have 3 explanations generated by LLMs, $j\in[1,3]$. Corresponding to each situation \(c\), we associate the expert-annotated explanation \(h_c\), and for each LLM response \(j\), the model-generated explanation \(e_j\). We calculate the similarity between \(h_c\) and \(e_j\) via the metric \(\text{sim}(h_c, e_j)\), with higher metric values indicating increased similarity. These similarity metrics are treated as unnormalized log probabilities (logits), to which we apply a softmax transformation for the derivation of probability values:

$$p_{cj} = \frac{\exp(\text{sim}(h_c, e_j))}{\sum_{c=1}^{3} \exp(\text{sim}(h_c, e_k))}$$

Subsequently, we impose the cross-entropy loss on the probabilities \(p_{cj}\) to yield the matching score:

$$\text{Matching}_{j} = -\sum_{c=1}^{3} y_{c}\log(p_{cj})$$

In this context, \(y_{c}\) is assigned a value of 1 when the expert-annotated explanation \(h_c\) corresponds with the LLM-generated explanation \(e_j\) (i.e., when \(c=j\)), signifying a perfect match. In contrast, \(y_{c}\) assumes a value of 0 for non-matching explanations. With cross-entropy loss and our 3-situation-designed dataset, $\text{Matching}_{j}$ in the equation quantitatively evaluate whether the LLM-generated explanation matches to the situation. Moreover, the cross-entropy loss $\text{Matching}_{j}$ is minimized when the LLM-generated explanation matches the situation. Compared with the similarity score, the matching score enables a quantitative assessment of the LLM's explanation adequacy. 

Ultimately, the similarity score evaluates the degree to which explanations generated by the LLM align with human-annotated references, while the matching score quantifies the appropriateness of these explanations in their specific situational contexts.

\section{Results}

Our results demonstrate several key findings regarding the efficacy of different prompting techniques in generating situated NLEs. Specific results are provided in the appendix; Figure \ref{sim_detail}, \ref{mat_detail}.

\subsection{How do prompt techniques matter?}
\begin{figure*}[t]
\centering 
\includegraphics[width=16cm]{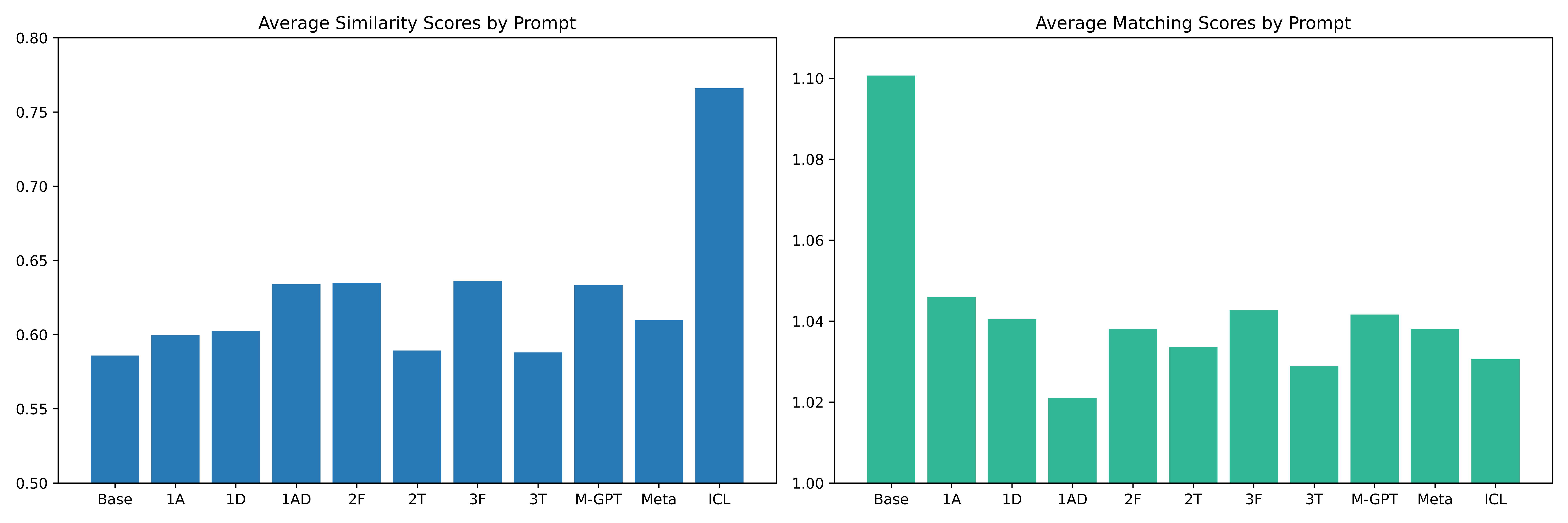} 
\caption{Average similarity and matching scores for all prompt techniques. 'M-GPT' refers to the use of GPT-3.5-turbo to generate prompts for situated NLE. 'Meta' refers to using the response model itself to generate prompts and respond to those. Note: A decrease in the matching score correlates with an enhancement in model performance on situated NLE tasks.} 
\label{image2}
\end{figure*}

Figure \ref{image2} shows the performance of each prompt techniques. Similarity score and matching score of 1A in the figure is the average score of explanations generated by prompt 1A2F3F, 1A2T3F, 1A2F3T, 1A2T3T and 7 LLMs(GPT-4-turbo, Gemini-pro, GPT-3.5-turbo, Pythia-2.8B, LLaMa2-7B, LLaMa2-13B-chat, and Yi-34B).

\paragraph{Specify the audience or the desired feature}
Specifying both the audience and the desired feature(1AD) can lead LLMs to generate more suitable explanations comparing with only specify the audience(1A) or only specify the desired feature(1D). The technique specifying both audience and desired feature (1AD) yielded the best results, with an average similarity score of 0.634 and a matching score of 1.021, indicating that providing comprehensive contextual information significantly enhances model performance. Comparatively, specifying either the desire feature (1D) performs slightly better than specifying the desired feature (1A)(average similarity scores of 0.602 and 0.599, respectively, and matching scores of 1.040 and 1.046). This suggests that while each element alone provides some contextual grounding, their combination is more potent in guiding the model to generate relevant and precise explanations. Thus, we recommend to specify both the audience and the desired feature for a situated explanation.

\paragraph{Adopt a persona}
The results show that the do not simulate the model as a helpful assistant (2F) approach yields a higher similarity score (0.635) compared to simulating the model as a helpful assistant (2T) (0.590), indicating that explanations generated without the persona are closer to human-annotated explanations. However, the matching score for 2T (1.034) is slightly better than for 2F (1.038), suggesting a marginally better alignment with the situational context when a persona is adopted. Despite this, the performance difference in matching scores is minimal, indicating that adding "You are a helpful assistant" to prompts for situated NLE tasks does not significantly aid the model's inferencing capability. Therefore, employing a persona in prompts is optional and may not be necessary for effective situated NLE generation.


\paragraph{Elicit the NLE with complete sentences}
The results highlight that use because (3F) method achieves a superior similarity score (0.6361) compared to use a complete sentence (3T) (0.5880), demonstrating that incorporating "because" in prompts helps the language model generate explanations that are significantly closer to those in \textsc{SBE}. Although 3T achieves a marginally better matching score (1.0289) than 3F (1.0427), this improvement is not substantial. Given the clearer advantage in similarity scores with 3F, we recommend using "because" in prompts for situated NLE tasks to more effectively align the generated explanations with the human-annotated standards.

\paragraph{Meta prompt}
In our exploration of meta prompts, we observed that these did not perform as well as those generated through the 1AD method. A notable issue with meta prompts is their tendency to include additional, often unnecessary specifications that may not align with the situational needs. For example, meta prompts including “Discuss the potential consequences of this problem and the importance of addressing it” introduce requirements that might not be relevant for the user. While maintaining fairness in our experimental evaluations, such specifics included in meta prompts led to their underperformance. This suggests that despite the innovative approach of using meta prompts, the traditional 1AD method remains more effective for generating situated natural language explanations aligned with the specific user contexts.

\paragraph{In-context learning prompt}
The performance of in-context learning prompts with regard to similarity scores is exemplary, demonstrating a state-of-the-art capability to replicate human-annotated explanations. Nonetheless, the performance in matching scores suggests that these prompts may not effectively aid the model in comprehending the situational context. This indicates a potential area for further refinement to enhance the model's situational awareness and its ability to generate contextually appropriate responses.

\begin{table}[ht]
\centering
\resizebox{0.7\columnwidth}{!}{%
\begin{tabular}{lrr}
\hline
                    & \textbf{Similarity} & \textbf{Matching} \\ \hline
Avg.                & 0.625               & 1.042             \\ \hline
ICL                 & 0.766               & 1.031             \\ \hline
\end{tabular}
}
\caption{Performance of the in-context learning prompt technique compared with the average performance on all prompt techniques.}
\end{table}

\subsection{How do different LLMs perform?}
\begin{figure*}[t]
\centering 
\includegraphics[width=.8\linewidth]{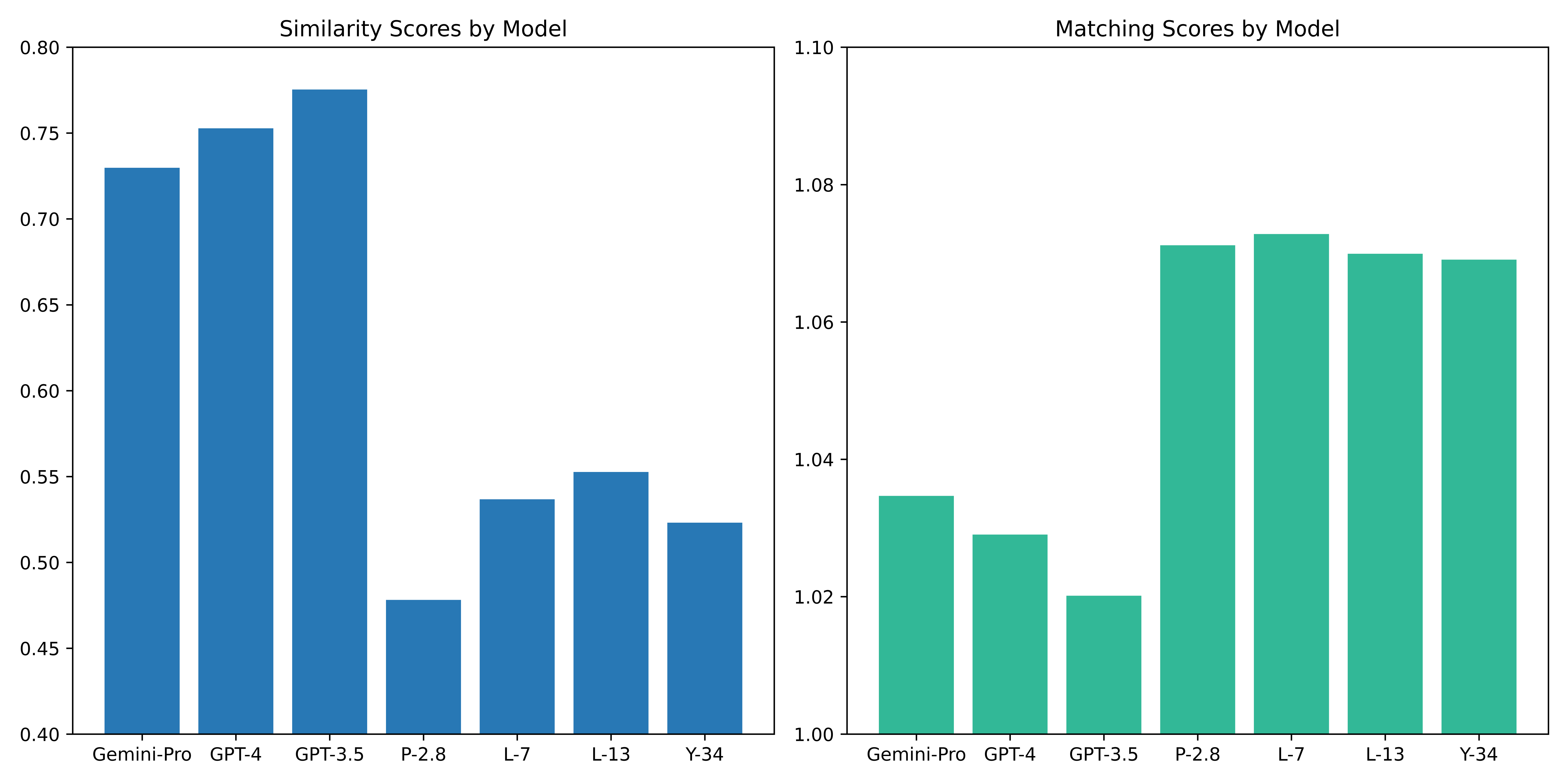} 
\caption{Average similarity and matching scores for all LLMs: 'P-2.8' represents Pythia-2.8B, 'L-7' stands for LLaMa-7B, 'L-13' is LLaMa-13B, and 'Y-34' indicates Yi-34B. Note: A decrease in the matching score correlates with an enhancement in model performance on situated NLE tasks.} 
\label{image3}
\end{figure*}

As Figure \ref{image3} shows, the performance varies by LLM.

\paragraph{Why are GPT-4 and Gemini-Pro worse than GPT-3.5?}
Why are GPT-4 and Gemini-Pro considered worse than GPT-3.5 in our evaluation? The explanations generated by GPT-4 and Gemini-Pro tend to be overly detailed, usually presented in a list format with up to seven points. (Appendix; Tabel \ref{case_GPT-4}, \ref{case_gemini-pro}) This makes their explanations not only too lengthy but also excessively specific. For instance, the average length of a human-annotated explanation is about 29.5 words, while the average for GPT-4-generated explanations is 421.9 words, and for Gemini-Pro, it's 198.7 words. In contrast, GPT-3.5-generated explanations have an average length of 137.6 words. Consequently, the performance of GPT-4 and Gemini-Pro is deemed inferior under our evaluation metrics, as their lengthy and overly specific outputs do not align well with our human-annotated explanations.

\paragraph{Commercial vs. open-source LLMs}
In our comparisons, models such as GPT-4, GPT-3.5, and Gemini-Pro, which are developed with significant commercial backing, consistently outperform their open-source counterparts. These commercially-developed models excel in generating more suitable, contextually appropriate explanations.

\paragraph{Variations among open-source LLMs}
When examining open-source language models, particularly those within the same family or architecture, we observe that the performance differences are not as stark in terms of matching scores. However, in terms of similarity scores, there is a clear hierarchy: Pythia-2.8B < Yi-34B < LlaMa-7B < LlaMa-13B. Although larger models generally show enhanced capabilities, suggesting that size contributes to model effectiveness, it is not the only factor influencing performance. Notably, LlaMa-7B and LlaMa-13B outperforms the larger Yi-34B model, indicating that factors beyond mere scale, such as model design, training protocol, or data quality, also play critical roles in determining a model's effectiveness. One the other hand, LlaMa-13B achieves better performance than LlaMa-13B. This result reinforces the idea that within a consistent architectural and training framework, larger models tend to demonstrate superior capabilities.

\section{Discussion}

\paragraph{Variance in Scenario Performance}
Our analysis reveals an interesting pattern in the variance of model outputs across different types of scenarios. Contrary to expectations, we find more variance among model outputs in scenarios related to daily life, which include a wide range of everyday activities and social interactions. This greater variance could be attributed to the inherent complexity and variability of such situations in daily life, which are less standardized and thus more challenging to simulate accurately. By contrast, scenarios involving specialized domains such as technology, politics, and health exhibited less variance among the outputs from different models. These areas often involve more standardized and well-defined concepts and terminologies, which are easier for models to inference. As a result, language models appear to handle these topics with greater consistency, possibly due to the clearer and more uniform contexts provided in such scenarios.

\section{Conclusion}
Based on the task of situated natural language explanations, this paper introduces \textsc{SBE}, a novel benchmarking dataset including audiences with desired features in specific situations. We use quantitative methods to evaluate different prompts and different LLMs and we rigorously evaluate the effectiveness of various prompting techniques and the performance of diverse large language models using quantitative methods. Our findings not only demonstrate the strengths and limitations of current approaches but also prepare future research to enhance the adaptability and precision of automated explanations tailored to distinct user contexts.

\section*{Limitations}
While our study marks a significant advancement in the field of situated natural language explanations (NLEs) by introducing the \textsc{SBE} and demonstrating the adaptability of language models to various contexts, it inherently simplifies the complex reality of potential real-life situations. \textsc{SBE}, designed with three specific contexts per explanandum, offers a quantitative approach to evaluating model performance but does not encompass the near-infinite variety of scenarios shaped by diverse audiences and their unique needs. Consequently, the results, though robust within the defined parameters, may not fully capture model effectiveness in more dynamically varied or extensively nuanced real-world applications. Future research should focus on expanding the dataset to cover a broader spectrum of situations among different backgrounds and biases. Moreover, refining the models to enhance their adaptability to the multifaceted nature of real-life contexts should be considered.

\section*{Ethics Statement}
In our research, we simulate hypothetical user situations to generate tailored explanations, which might, on LLM agents, trigger the use of personal data if not deployed properly. The implication of deploying similar technology in real-world settings raises significant privacy concerns. Although our study does not entail these risks due to the nature of our data, future LLM agents must consider implementing stringent data protection measures. These should include robust anonymization techniques, minimal data retention policies, and adherence to privacy regulations to safeguard individual data rights. 
Additionally, as LLM-generated explanations become more convincing, they could be adversarially helpful \citep{ajwani2024llmgenerated}. Approaches to defend adversarial helpfulness include asking the explainers to present information from multiple perspectives, which is relevant to adaptation.
Last but not least, it is crucial to clarify that the explanations generated by our models are algorithmic outputs and do not reflect personal beliefs or empirical truths.


\bibliography{custom}

\begin{thebibliography}{41}
\expandafter\ifx\csname natexlab\endcsname\relax\def\natexlab#1{#1}\fi

\bibitem[{Administration(2011)}]{administrationWrite}
General~Services Administration. 2011.
\newblock \href {https://www.plainlanguage.gov/guidelines/audience/} {Write for your audience}.

\bibitem[{Aggarwal et~al.(2021)Aggarwal, Mandowara, Agrawal, Khandelwal, Singla, and Garg}]{aggarwal-etal-2021-explanations}
Shourya Aggarwal, Divyanshu Mandowara, Vishwajeet Agrawal, Dinesh Khandelwal, Parag Singla, and Dinesh Garg. 2021.
\newblock \href {https://doi.org/10.18653/v1/2021.acl-long.238} {{E}xplanations for {C}ommonsense{QA}: {N}ew {D}ataset and {M}odels}.
\newblock In \emph{Proceedings of the 59th Annual Meeting of the Association for Computational Linguistics and the 11th International Joint Conference on Natural Language Processing (Volume 1: Long Papers)}, pages 3050--3065, Online. Association for Computational Linguistics.

\bibitem[{AI et~al.(2024)AI, :, Young, Chen, Li, Huang, Zhang, Zhang, Li, Zhu, Chen, Chang, Yu, Liu, Liu, Yue, Yang, Yang, Yu, Xie, Huang, Hu, Ren, Niu, Nie, Xu, Liu, Wang, Cai, Gu, Liu, and Dai}]{ai2024yi}
01. AI, :, Alex Young, Bei Chen, Chao Li, Chengen Huang, Ge~Zhang, Guanwei Zhang, Heng Li, Jiangcheng Zhu, Jianqun Chen, Jing Chang, Kaidong Yu, Peng Liu, Qiang Liu, Shawn Yue, Senbin Yang, Shiming Yang, Tao Yu, Wen Xie, Wenhao Huang, Xiaohui Hu, Xiaoyi Ren, Xinyao Niu, Pengcheng Nie, Yuchi Xu, Yudong Liu, Yue Wang, Yuxuan Cai, Zhenyu Gu, Zhiyuan Liu, and Zonghong Dai. 2024.
\newblock \href {http://arxiv.org/abs/2403.04652} {Yi: Open foundation models by 01.ai}.

\bibitem[{Ajwani et~al.(2024)Ajwani, Javaji, Rudzicz, and Zhu}]{ajwani2024llmgenerated}
Rohan Ajwani, Shashidhar~Reddy Javaji, Frank Rudzicz, and Zining Zhu. 2024.
\newblock \href {http://arxiv.org/abs/2405.06800} {{LLM-Generated Black-box Explanations Can Be Adversarially Helpful}}.

\bibitem[{Biderman et~al.(2023)Biderman, Schoelkopf, Anthony, Bradley, O'Brien, Hallahan, Khan, Purohit, Prashanth, Raff, Skowron, Sutawika, and Van Der~Wal}]{10.5555/3618408.3618510}
Stella Biderman, Hailey Schoelkopf, Quentin Anthony, Herbie Bradley, Kyle O'Brien, Eric Hallahan, Mohammad~Aflah Khan, Shivanshu Purohit, USVSN~Sai Prashanth, Edward Raff, Aviya Skowron, Lintang Sutawika, and Oskar Van Der~Wal. 2023.
\newblock Pythia: a suite for analyzing large language models across training and scaling.
\newblock In \emph{Proceedings of the 40th International Conference on Machine Learning}, ICML'23. JMLR.org.

\bibitem[{Boyd-Graber et~al.(2022)Boyd-Graber, Carton, Feng, Liao, Lombrozo, Smith-Renner, and Tan}]{boyd-graber-etal-2022-human}
Jordan Boyd-Graber, Samuel Carton, Shi Feng, Q.~Vera Liao, Tania Lombrozo, Alison Smith-Renner, and Chenhao Tan. 2022.
\newblock \href {https://doi.org/10.18653/v1/2022.naacl-tutorials.4} {Human-centered evaluation of explanations}.
\newblock In \emph{Proceedings of the 2022 Conference of the North American Chapter of the Association for Computational Linguistics: Human Language Technologies: Tutorial Abstracts}, pages 26--32, Seattle, United States. Association for Computational Linguistics.

\bibitem[{Camburu et~al.(2020)Camburu, Shillingford, Minervini, Lukasiewicz, and Blunsom}]{camburu-etal-2020-make}
Oana-Maria Camburu, Brendan Shillingford, Pasquale Minervini, Thomas Lukasiewicz, and Phil Blunsom. 2020.
\newblock \href {https://doi.org/10.18653/v1/2020.acl-main.382} {Make up your mind! adversarial generation of inconsistent natural language explanations}.
\newblock In \emph{Proceedings of the 58th Annual Meeting of the Association for Computational Linguistics}, pages 4157--4165, Online. Association for Computational Linguistics.

\bibitem[{Chan et~al.(2022)Chan, Nie, Tan, Peng, Firooz, Sanjabi, and Ren}]{chan2022frame}
Aaron Chan, Shaoliang Nie, Liang Tan, Xiaochang Peng, Hamed Firooz, Maziar Sanjabi, and Xiang Ren. 2022.
\newblock Frame: Evaluating rationale-label consistency metrics for free-text rationales.
\newblock \emph{arXiv preprint arXiv:2207.00779}.

\bibitem[{Chen et~al.(2023)Chen, Brahman, Ren, Ji, Choi, and Swayamdipta}]{chen-etal-2023-rev}
Hanjie Chen, Faeze Brahman, Xiang Ren, Yangfeng Ji, Yejin Choi, and Swabha Swayamdipta. 2023.
\newblock \href {https://doi.org/10.18653/v1/2023.acl-long.112} {{REV}: Information-theoretic evaluation of free-text rationales}.
\newblock In \emph{Proceedings of the 61st Annual Meeting of the Association for Computational Linguistics (Volume 1: Long Papers)}, pages 2007--2030, Toronto, Canada. Association for Computational Linguistics.

\bibitem[{Clinciu et~al.(2021)Clinciu, Eshghi, and Hastie}]{clinciu-etal-2021-study}
Miruna-Adriana Clinciu, Arash Eshghi, and Helen Hastie. 2021.
\newblock \href {https://doi.org/10.18653/v1/2021.eacl-main.202} {A study of automatic metrics for the evaluation of natural language explanations}.
\newblock In \emph{Proceedings of the 16th Conference of the European Chapter of the Association for Computational Linguistics: Main Volume}, pages 2376--2387, Online. Association for Computational Linguistics.

\bibitem[{Cutts(2020)}]{cutts2020oxford}
Martin Cutts. 2020.
\newblock \emph{Oxford guide to plain English}.
\newblock Oxford University Press, USA.

\bibitem[{DeYoung et~al.(2020)DeYoung, Jain, Rajani, Lehman, Xiong, Socher, and Wallace}]{deyoung-etal-2020-eraser}
Jay DeYoung, Sarthak Jain, Nazneen~Fatema Rajani, Eric Lehman, Caiming Xiong, Richard Socher, and Byron~C. Wallace. 2020.
\newblock \href {https://doi.org/10.18653/v1/2020.acl-main.408} {{ERASER}: {A} benchmark to evaluate rationalized {NLP} models}.
\newblock In \emph{Proceedings of the 58th Annual Meeting of the Association for Computational Linguistics}, pages 4443--4458, Online. Association for Computational Linguistics.

\bibitem[{Ehsan et~al.(2024)Ehsan, Passi, Liao, Chan, Lee, Muller, and Riedl}]{ehsan2024Who}
Upol Ehsan, Samir Passi, Q.~Vera Liao, Larry Chan, I.-Hsiang Lee, Michael Muller, and Mark~O. Riedl. 2024.
\newblock \href {http://arxiv.org/abs/2107.13509} {The {{Who}} in {{XAI}}: {{How AI Background Shapes Perceptions}} of {{AI Explanations}}}.
\newblock In \emph{Proceedings of the {{CHI Conference}} on {{Human Factors}} in {{Computing Systems}}}, pages 1--32.

\bibitem[{Geng et~al.(2022{\natexlab{a}})Geng, Fu, Ge, Li, {de Melo}, and Zhang}]{geng2022Improving}
Shijie Geng, Zuohui Fu, Yingqiang Ge, Lei Li, Gerard {de Melo}, and Yongfeng Zhang. 2022{\natexlab{a}}.
\newblock \href {https://aclanthology.org/2022.acl-long.20} {Improving {{Personalized Explanation Generation}} through {{Visualization}}}.
\newblock In \emph{Proceedings of the 60th {{Annual Meeting}} of the {{Association}} for {{Computational Linguistics}} ({{Volume}} 1: {{Long Papers}})}, pages 244--255, Dublin, Ireland. Association for Computational Linguistics.

\bibitem[{Geng et~al.(2022{\natexlab{b}})Geng, Liu, Fu, Ge, and Zhang}]{geng2022recommendation}
Shijie Geng, Shuchang Liu, Zuohui Fu, Yingqiang Ge, and Yongfeng Zhang. 2022{\natexlab{b}}.
\newblock Recommendation as language processing (rlp): A unified pretrain, personalized prompt \& predict paradigm (p5).
\newblock In \emph{Proceedings of the 16th ACM Conference on Recommender Systems}, pages 299--315.

\bibitem[{Goyal et~al.(2023)Goyal, Briakou, Liu, Baumler, Bonial, Micher, Voss, Carpuat, and Daum{\'e}~III}]{goyal2023What}
Navita Goyal, Eleftheria Briakou, Amanda Liu, Connor Baumler, Claire Bonial, Jeffrey Micher, Clare Voss, Marine Carpuat, and Hal Daum{\'e}~III. 2023.
\newblock \href {https://aclanthology.org/2023.emnlp-main.201} {What {{Else Do I Need}} to {{Know}}? {{The Effect}} of {{Background Information}} on {{Users}}' {{Reliance}} on {{QA Systems}}}.
\newblock In \emph{Proceedings of the 2023 {{Conference}} on {{Empirical Methods}} in {{Natural Language Processing}}}, pages 3313--3330, Singapore. Association for Computational Linguistics.

\bibitem[{Joshi et~al.(2023)Joshi, Liu, Ramnath, Chan, Tong, Nie, Wang, Choi, and Ren}]{joshi-etal-2023-machine}
Brihi Joshi, Ziyi Liu, Sahana Ramnath, Aaron Chan, Zhewei Tong, Shaoliang Nie, Qifan Wang, Yejin Choi, and Xiang Ren. 2023.
\newblock \href {https://doi.org/10.18653/v1/2023.acl-long.392} {Are machine rationales (not) useful to humans? measuring and improving human utility of free-text rationales}.
\newblock In \emph{Proceedings of the 61st Annual Meeting of the Association for Computational Linguistics (Volume 1: Long Papers)}, pages 7103--7128, Toronto, Canada. Association for Computational Linguistics.

\bibitem[{Li et~al.(2020)Li, Zhang, and Chen}]{10.1145/3340531.3411992}
Lei Li, Yongfeng Zhang, and Li~Chen. 2020.
\newblock \href {https://doi.org/10.1145/3340531.3411992} {Generate neural template explanations for recommendation}.
\newblock In \emph{Proceedings of the 29th ACM International Conference on Information \& Knowledge Management}, CIKM '20, page 755–764, New York, NY, USA. Association for Computing Machinery.

\bibitem[{Liao et~al.(2022)Liao, Zhang, Luss, Doshi-Velez, and Dhurandhar}]{Liao_Zhang_Luss_Doshi-Velez_Dhurandhar_2022}
Q.~Vera Liao, Yunfeng Zhang, Ronny Luss, Finale Doshi-Velez, and Amit Dhurandhar. 2022.
\newblock \href {https://doi.org/10.1609/hcomp.v10i1.21995} {Connecting algorithmic research and usage contexts: A perspective of contextualized evaluation for explainable ai}.
\newblock \emph{Proceedings of the AAAI Conference on Human Computation and Crowdsourcing}, 10(1):147--159.

\bibitem[{Liu et~al.(2024)Liu, Diab, and Fried}]{liu2024Evaluating}
Andy Liu, Mona Diab, and Daniel Fried. 2024.
\newblock \href {http://arxiv.org/abs/2405.20253} {Evaluating large language model biases in persona-steered generation}.
\newblock In \emph{Proceedings of the 2024 Annual Conference of the Association for Computational Linguistics}.

\bibitem[{Lombrozo(2007)}]{LOMBROZO2007232}
Tania Lombrozo. 2007.
\newblock \href {https://doi.org/https://doi.org/10.1016/j.cogpsych.2006.09.006} {Simplicity and probability in causal explanation}.
\newblock \emph{Cognitive Psychology}, 55(3):232--257.

\bibitem[{Lyu et~al.(2023)Lyu, Havaldar, Stein, Zhang, Rao, Wong, Apidianaki, and Callison-Burch}]{lyu-etal-2023-faithful}
Qing Lyu, Shreya Havaldar, Adam Stein, Li~Zhang, Delip Rao, Eric Wong, Marianna Apidianaki, and Chris Callison-Burch. 2023.
\newblock \href {https://doi.org/10.18653/v1/2023.ijcnlp-main.20} {Faithful chain-of-thought reasoning}.
\newblock In \emph{Proceedings of the 13th International Joint Conference on Natural Language Processing and the 3rd Conference of the Asia-Pacific Chapter of the Association for Computational Linguistics (Volume 1: Long Papers)}, pages 305--329, Nusa Dua, Bali. Association for Computational Linguistics.

\bibitem[{Marasovic et~al.(2022)Marasovic, Beltagy, Downey, and Peters}]{marasovic-etal-2022-shot}
Ana Marasovic, Iz~Beltagy, Doug Downey, and Matthew Peters. 2022.
\newblock \href {https://doi.org/10.18653/v1/2022.findings-naacl.31} {Few-shot self-rationalization with natural language prompts}.
\newblock In \emph{Findings of the Association for Computational Linguistics: NAACL 2022}, pages 410--424, Seattle, United States. Association for Computational Linguistics.

\bibitem[{Miller(2019)}]{MILLER20191}
Tim Miller. 2019.
\newblock \href {https://doi.org/https://doi.org/10.1016/j.artint.2018.07.007} {Explanation in artificial intelligence: Insights from the social sciences}.
\newblock \emph{Artificial Intelligence}, 267:1--38.

\bibitem[{Narang et~al.(2020)Narang, Raffel, Lee, Roberts, Fiedel, and Malkan}]{narang2020wt5}
Sharan Narang, Colin Raffel, Katherine Lee, Adam Roberts, Noah Fiedel, and Karishma Malkan. 2020.
\newblock Wt5?! training text-to-text models to explain their predictions.
\newblock \emph{arXiv preprint arXiv:2004.14546}.

\bibitem[{Ouyang et~al.(2022)Ouyang, Wu, Jiang, Almeida, Wainwright, Mishkin, Zhang, Agarwal, Slama, Ray, Schulman, Hilton, Kelton, Miller, Simens, Askell, Welinder, Christiano, Leike, and Lowe}]{NEURIPS2022_b1efde53}
Long Ouyang, Jeffrey Wu, Xu~Jiang, Diogo Almeida, Carroll Wainwright, Pamela Mishkin, Chong Zhang, Sandhini Agarwal, Katarina Slama, Alex Ray, John Schulman, Jacob Hilton, Fraser Kelton, Luke Miller, Maddie Simens, Amanda Askell, Peter Welinder, Paul~F Christiano, Jan Leike, and Ryan Lowe. 2022.
\newblock \href {https://proceedings.neurips.cc/paper_files/paper/2022/file/b1efde53be364a73914f58805a001731-Paper-Conference.pdf} {Training language models to follow instructions with human feedback}.
\newblock In \emph{Advances in Neural Information Processing Systems}, volume~35, pages 27730--27744. Curran Associates, Inc.

\bibitem[{Purdue()}]{purdueTone}
Online Writing~Lab Purdue.
\newblock \href {https://owl.purdue.edu/owl/general_writing/writing_style/diction/tone_mood_audience.html} {Tone, {{Mood}}, and {{Audience}} - {{Purdue OWL}}{\textregistered} - {{Purdue University}}}.

\bibitem[{Rajani et~al.(2019)Rajani, McCann, Xiong, and Socher}]{rajani-etal-2019-explain}
Nazneen~Fatema Rajani, Bryan McCann, Caiming Xiong, and Richard Socher. 2019.
\newblock \href {https://doi.org/10.18653/v1/P19-1487} {Explain yourself! leveraging language models for commonsense reasoning}.
\newblock In \emph{Proceedings of the 57th Annual Meeting of the Association for Computational Linguistics}, pages 4932--4942, Florence, Italy. Association for Computational Linguistics.

\bibitem[{Rao et~al.(2024)Rao, Yerukola, Shah, Reinecke, and Sap}]{rao2024normad}
Abhinav Rao, Akhila Yerukola, Vishwa Shah, Katharina Reinecke, and Maarten Sap. 2024.
\newblock Normad: A benchmark for measuring the cultural adaptability of large language models.
\newblock \emph{arXiv preprint arXiv:2404.12464}.

\bibitem[{Reimers and Gurevych(2019{\natexlab{a}})}]{reimers-gurevych-2019-sentence}
Nils Reimers and Iryna Gurevych. 2019{\natexlab{a}}.
\newblock \href {https://doi.org/10.18653/v1/D19-1410} {Sentence-{BERT}: Sentence embeddings using {S}iamese {BERT}-networks}.
\newblock In \emph{Proceedings of the 2019 Conference on Empirical Methods in Natural Language Processing and the 9th International Joint Conference on Natural Language Processing (EMNLP-IJCNLP)}, pages 3982--3992, Hong Kong, China. Association for Computational Linguistics.

\bibitem[{Reimers and Gurevych(2019{\natexlab{b}})}]{reimers-2019-sentence-bert}
Nils Reimers and Iryna Gurevych. 2019{\natexlab{b}}.
\newblock \href {http://arxiv.org/abs/1908.10084} {Sentence-bert: Sentence embeddings using siamese bert-networks}.
\newblock In \emph{Proceedings of the 2019 Conference on Empirical Methods in Natural Language Processing}. Association for Computational Linguistics.

\bibitem[{Si et~al.(2024)Si, Goyal, Wu, Zhao, Feng, Daum{\'e}~III, and Boyd-Graber}]{si2024large}
Chenglei Si, Navita Goyal, Sherry~Tongshuang Wu, Chen Zhao, Shi Feng, Hal Daum{\'e}~III, and Jordan Boyd-Graber. 2024.
\newblock Large language models help humans verify truthfulness--except when they are convincingly wrong.
\newblock \emph{NAACL}.

\bibitem[{Stephen et~al.(2022)Stephen, Kiefer, Kowalski, and Bennett}]{stephen2022Guide}
Reid Stephen, Kate Kiefer, Dawn Kowalski, and Andrea Bennett. 2022.
\newblock \href {https://writing.colostate.edu/guides/guide.cfm?guideid=19} {Guide: {{Adapting}} to {{Your Audience}}}.

\bibitem[{Touvron et~al.(2023)Touvron, Martin, Stone, Albert, Almahairi, Babaei, Bashlykov, Batra, Bhargava, Bhosale, Bikel, Blecher, Ferrer, Chen, Cucurull, Esiobu, Fernandes, Fu, Fu, Fuller, Gao, Goswami, Goyal, Hartshorn, Hosseini, Hou, Inan, Kardas, Kerkez, Khabsa, Kloumann, Korenev, Koura, Lachaux, Lavril, Lee, Liskovich, Lu, Mao, Martinet, Mihaylov, Mishra, Molybog, Nie, Poulton, Reizenstein, Rungta, Saladi, Schelten, Silva, Smith, Subramanian, Tan, Tang, Taylor, Williams, Kuan, Xu, Yan, Zarov, Zhang, Fan, Kambadur, Narang, Rodriguez, Stojnic, Edunov, and Scialom}]{touvron2023LLaMa}
Hugo Touvron, Louis Martin, Kevin Stone, Peter Albert, Amjad Almahairi, Yasmine Babaei, Nikolay Bashlykov, Soumya Batra, Prajjwal Bhargava, Shruti Bhosale, Dan Bikel, Lukas Blecher, Cristian~Canton Ferrer, Moya Chen, Guillem Cucurull, David Esiobu, Jude Fernandes, Jeremy Fu, Wenyin Fu, Brian Fuller, Cynthia Gao, Vedanuj Goswami, Naman Goyal, Anthony Hartshorn, Saghar Hosseini, Rui Hou, Hakan Inan, Marcin Kardas, Viktor Kerkez, Madian Khabsa, Isabel Kloumann, Artem Korenev, Punit~Singh Koura, Marie-Anne Lachaux, Thibaut Lavril, Jenya Lee, Diana Liskovich, Yinghai Lu, Yuning Mao, Xavier Martinet, Todor Mihaylov, Pushkar Mishra, Igor Molybog, Yixin Nie, Andrew Poulton, Jeremy Reizenstein, Rashi Rungta, Kalyan Saladi, Alan Schelten, Ruan Silva, Eric~Michael Smith, Ranjan Subramanian, Xiaoqing~Ellen Tan, Binh Tang, Ross Taylor, Adina Williams, Jian~Xiang Kuan, Puxin Xu, Zheng Yan, Iliyan Zarov, Yuchen Zhang, Angela Fan, Melanie Kambadur, Sharan Narang, Aurelien Rodriguez, Robert Stojnic, Sergey Edunov, and Thomas
  Scialom. 2023.
\newblock \href {http://arxiv.org/abs/2307.09288} {Llama 2: Open foundation and fine-tuned chat models}.

\bibitem[{{van Merri{\"e}nboer} and Sweller(2005)}]{vanmerrienboer2005Cognitive}
Jeroen J.~G. {van Merri{\"e}nboer} and John Sweller. 2005.
\newblock \href {https://doi.org/10.1007/s10648-005-3951-0} {Cognitive {{Load Theory}} and {{Complex Learning}}: {{Recent Developments}} and {{Future Directions}}}.
\newblock \emph{Educ Psychol Rev}, 17(2):147--177.

\bibitem[{Wang et~al.(2024)Wang, Yu, Zhang, Qi, Sap, Neubig, Bisk, and Zhu}]{wang2024sotopia}
Ruiyi Wang, Haofei Yu, Wenxin Zhang, Zhengyang Qi, Maarten Sap, Graham Neubig, Yonatan Bisk, and Hao Zhu. 2024.
\newblock Sotopia-$\pi$: Interactive learning of socially intelligent language agents.
\newblock \emph{arXiv preprint arXiv:2403.08715}.

\bibitem[{Wiegreffe et~al.(2022)Wiegreffe, Hessel, Swayamdipta, Riedl, and Choi}]{wiegreffe-etal-2022-reframing}
Sarah Wiegreffe, Jack Hessel, Swabha Swayamdipta, Mark Riedl, and Yejin Choi. 2022.
\newblock \href {https://doi.org/10.18653/v1/2022.naacl-main.47} {Reframing human-{AI} collaboration for generating free-text explanations}.
\newblock In \emph{Proceedings of the 2022 Conference of the North American Chapter of the Association for Computational Linguistics: Human Language Technologies}, pages 632--658, Seattle, United States. Association for Computational Linguistics.

\bibitem[{Wiegreffe and Marasovic(2021)}]{NEURIPSDatasetsBenchmarks2021}
Sarah Wiegreffe and Ana Marasovic. 2021.
\newblock \href {https://datasets-benchmarks-proceedings.neurips.cc/paper_files/paper/2021/file/698d51a19d8a121ce581499d7b701668-Paper.pdf} {Teach me to explain: A review of datasets for explainable natural language processing}.
\newblock In \emph{Proceedings of the Neural Information Processing Systems Track on Datasets and Benchmarks}, volume~1.

\bibitem[{Yerukola et~al.(2024)Yerukola, Vaduguru, Fried, and Sap}]{yerukola2024pope}
Akhila Yerukola, Saujas Vaduguru, Daniel Fried, and Maarten Sap. 2024.
\newblock Is the pope catholic? yes, the pope is catholic. generative evaluation of intent resolution in llms.
\newblock \emph{arXiv preprint arXiv:2405.08760}.

\bibitem[{Yeung et~al.(2020)Yeung, Joshi, Williams, and Rudzicz}]{yeung2020sequential}
Arnold~YS Yeung, Shalmali Joshi, Joseph~Jay Williams, and Frank Rudzicz. 2020.
\newblock \href {http://arxiv.org/abs/2007.09028} {Sequential explanations with mental model-based policies}.

\bibitem[{Zhu et~al.(2023)Zhu, Jiang, Yang, Nag, Zhang, Jie, Gao, Rudzicz, and Yin}]{zhu2023Situated}
Zining Zhu, Haoming Jiang, Jingfeng Yang, Sreyashi Nag, Chao Zhang, Huang Jie, Yifan Gao, Frank Rudzicz, and Bing Yin. 2023.
\newblock \href {https://arxiv.org/abs/2308.14115} {Situated {{Natural Languages Explanations}}}.
\newblock In \emph{{ACL NLRSE Workshop}}.

\end{thebibliography}
\bibliographystyle{acl_natbib}

\appendix

\section{All Prompt templates, Table \ref{tab:prompt_templates}}

\begin{table*}[hbp]
\centering
\scalebox{0.75}{
\begin{tabular}{ll}

\hline\textbf{Prompt Method} & \textbf{Prompt Content} \\
\hline
Base & \{explanandum\} because \\
\hline
1A2F3F & Following is an explanation towards \{audience\}. \{explanandum\} because \\
\hline
1A2F3T & Following is an explanation towards \{audience\}: \{explanandum\}. \\
\hline
1A2T3F & You are a helpful assistant explaining to \{audience\}. \{explanandum\} because \\
\hline
1A2T3T & You are a helpful assistant explaining to \{audience\}. \{explanandum\}. \\
\hline
1D2F3F & Following is an explanation about \{reason\}. \{explanandum\} because \\
\hline
1D2F3T & Following is an explanation about \{reason\}. \{explanandum\}. \\
\hline
1D2T3F & You are a helpful assistant explaining about \{desired feature\}. \{explanandum\} because \\
\hline
1D2T3T & You are a helpful assistant explaining about \{desired feature\}. \{explanandum\}. \\
\hline
1AD2F3F & Following is an explanation towards \{audience\}, about \{desired feature\}. \{explanandum\} because \\
\hline
1AD2F3T & Following is an explanation towards \{audience\}, about \{desired feature\}. \{explanandum\} \\
\hline
1AD2T3F & You are a helpful assistant explaining to \{audience\}, about \{desired feature\}. \{explanandum\} because \\
\hline
1AD2T3T & You are a helpful assistant explaining to \{audience\}, about \{desired feature\}. \{explanandum\} \\
\hline
Meta prompt & You are a helpful assistant helping me write a prompt. \\
&I want to write a prompt to generate an explanation about why \{explanandum\} to \{audience\}, about \{desired feature\}. \\
&Give me the prompt directly. \\
\hline
ICL prompt & "For audience\_1: \\
&Q: Following is an explanation towards \{audience2\}, about \{desired feature2\}. \{explanandum\} because \\
&A: \{explanation2\} \\
&Q: Following is an explanation towards \{audience3\}, about \{desired feature3\}. \{explanandum\} because \\
&A: \{explanation3\} \\
&Q: Following is an explanation towards \{audience1\}, about \{desired feature1\}. \{explanandum\} because \\
&A:" \\
\hline
\end{tabular}
}
\caption{All prompt templates used in the paper.}
\label{tab:prompt_templates}
\end{table*}

\section{Case Study, Table\ref{case_GPT-4}, \ref{case_gemini-pro}}
\begin{table*}
\small 
\begin{tabular}{|p{2cm}|p{14cm}|}
\hline
\textbf{User \newline instruction} & Following is an explanation towards students, about engaging and adaptive learning tools tailored to individual preferences. Educational technology can be meaningful because \\
\hline
\textbf{GPT-4 \newline generated \newline explanation} & Engaging and adaptive learning tools tailored to individual preferences are meaningful in educational technology because they help create personalized learning experiences that meet the unique needs and strengths of each student. Here are several reasons why these tools are particularly effective:

\begin{itemize}
  \item \textbf{Personalization}: Adaptive learning technologies can adjust the content, pace, and difficulty based on individual learning preferences and performance. This means students can learn in a way that is best suited to their own style and pace, which can help to foster deeper understanding and retention of knowledge.
  \item \textbf{Increased Engagement}: Engaging learning tools are designed to capture and maintain students' interest through interactive elements such as games, quizzes, and interactive simulations. When students are engaged, they are more likely to invest time and effort in their learning, which can lead to better outcomes.
  \item \textbf{Accessibility and Inclusivity}: Adaptive learning tools can also provide support for learners with diverse needs, including those with disabilities. For example, text can be converted to speech for students with visual impairments, or lessons can be structured in different ways to cater to various learning disabilities. This inclusivity ensures all students have an equal opportunity to learn and succeed.
  \item \textbf{Immediate Feedback}: These tools often provide immediate feedback, which is crucial for learning. Students can understand what they're doing right and where they need improvement. This immediate response helps them to adjust their strategies and efforts in real time, promoting a more efficient learning process.
  \item \textbf{Data-driven Insights}: Adaptive tools often collect data on student performance, which can be analyzed to provide insights into learning habits, difficulties, and progression. This information can be invaluable for teachers and educators to further tailor learning experiences, provide targeted support, and even refine curriculum and instructional strategies.
  \item \textbf{Scalability}: Technology enables these personalized learning experiences to be scaled across large numbers of students without the same increase in resources that would be necessary if done manually by educators. This is particularly valuable in under-resourced or rural areas.
  \item \textbf{Skill Preparation}: Modern educational tools prepare students for the technological world they will enter into after their education. By using tech-based tools in their learning process, students become proficient in digital literacy, a critical skill for almost all career paths in the modern workforce.
\end{itemize}
\\
\hline
\end{tabular}
\caption{An explanation generated by GPT-4.}
\label{case_GPT-4}
\end{table*}

\begin{table*}[htbp]
\small 
\begin{tabular}{|p{2cm}|p{14cm}|}
\hline
\textbf{User \newline instruction} & Following is an explanation towards students, about engaging and adaptive learning tools tailored to individual preferences. Educational technology can be meaningful because \\
\hline

\textbf{Gemini-Pro \newline generated \newline explanation} & \textbf{Engaging and Adaptive Learning Tools Tailored to Individual Preferences}

Educational technology can be transformative when it provides students with personalized learning experiences tailored to their unique needs and preferences. By using engaging and adaptive tools, educators can create environments where students can thrive and achieve their full potential.

\textbf{Engaging Tools:}

\begin{itemize}
  \item \textbf{Interactive Simulations and Games:} These provide immersive experiences that allow students to explore complex concepts in a fun and interactive way.
  \item \textbf{Virtual Reality (VR) Environments:} VR headsets transport students to virtual worlds, allowing them to experience real-life scenarios and conduct experiments in a safe and controlled environment.
  \item \textbf{Social Collaboration Tools:} Platforms that allow students to connect with peers, ask questions, and engage in group projects foster collaboration and enhance learning.
\end{itemize}

\textbf{Adaptive Tools:}

\begin{itemize}
  \item \textbf{Adaptive Learning Systems:} These adjust course content and assessments to match the individual student's learning pace, skill level, and learning style.
  \item \textbf{Intelligent Tutoring Systems (ITS):} ITSs provide personalized guidance and feedback to students, helping them identify areas for improvement and build on their strengths.
  \item \textbf{AI-Powered Language Tools:} These tools assist students with writing and reading, providing real-time feedback on grammar, vocabulary, and comprehension.
\end{itemize}

\textbf{Tailored to Individual Preferences:}

These tools can be tailored to cater to different learning preferences:

\begin{itemize}
  \item \textbf{Visual Learners:} Simulations, VR environments, and interactive videos engage visual learners.
  \item \textbf{Auditory Learners:} Audio-based lessons, podcasts, and recorded lectures cater to auditory learners.
  \item \textbf{Kinesthetic Learners:} Hands-on activities, simulations, and collaborative projects provide opportunities for kinesthetic learners.
\end{itemize}

\textbf{Meaningfulness of Educational Technology:}

By providing students with engaging and adaptive learning tools tailored to their individual preferences, educational technology can make learning:

\begin{itemize}
  \item \textbf{Personalized:} Content and assessments are tailored to meet the specific needs of each student.
  \item \textbf{Motivating:} Interactive and engaging tools keep students interested and motivated to learn.
  \item \textbf{Effective:} Adaptive systems track student progress and provide personalized feedback to enhance learning outcomes.
  \item \textbf{Equitable:} Technology provides all students with equal access to high-quality learning experiences, regardless of their background or learning style.
\end{itemize}
\\
\hline
\end{tabular}
\caption{An explanation generated by Gemini-Pro.}
\label{case_gemini-pro}
\end{table*}

\begin{figure*}[htbp]
\centering 
\includegraphics[width=\textwidth]{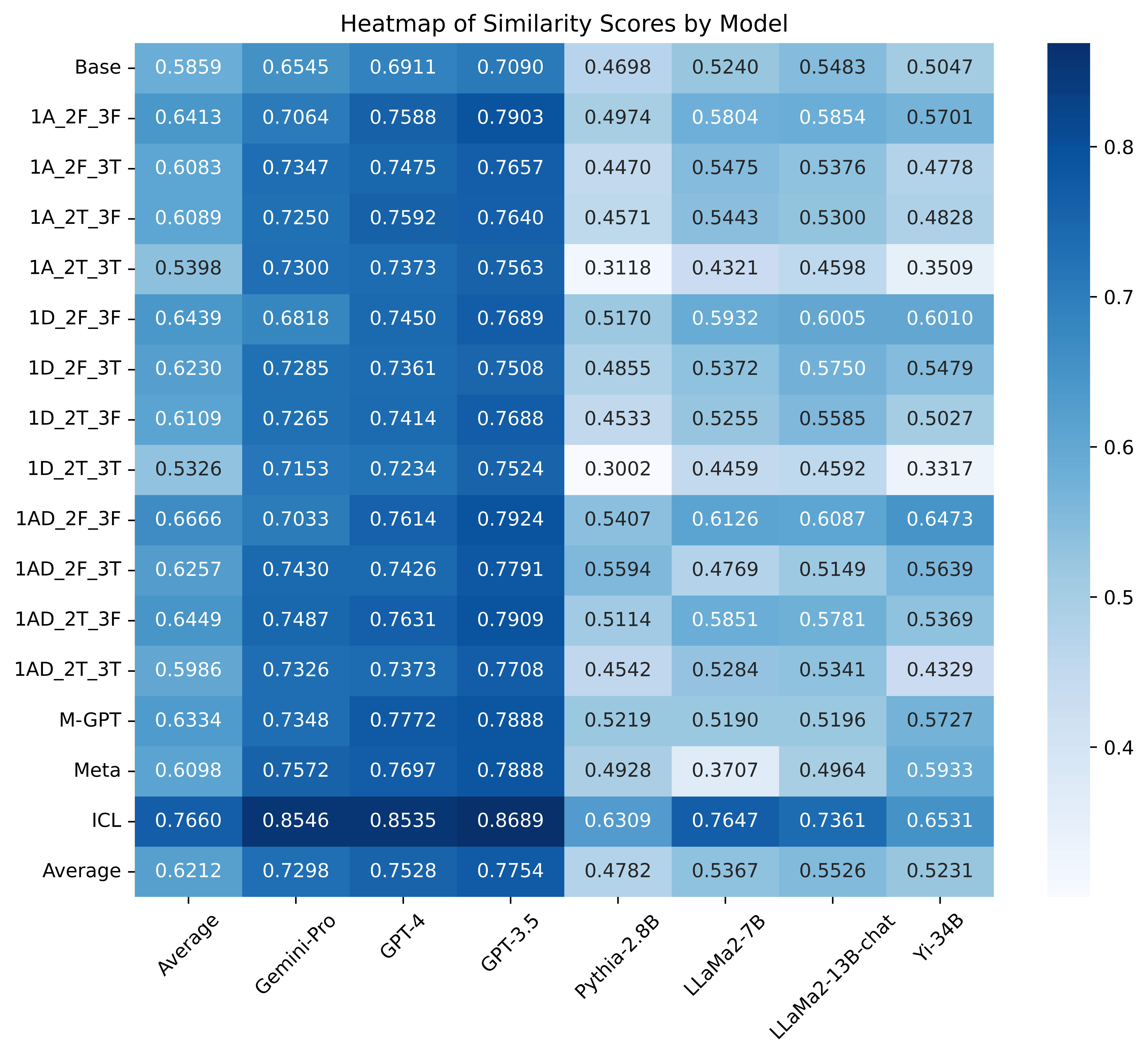} 
\caption{Similarity score heatmap.}
\label{sim_detail}
\end{figure*}

\begin{figure*}[htbp]
\centering 
\includegraphics[width=\textwidth]{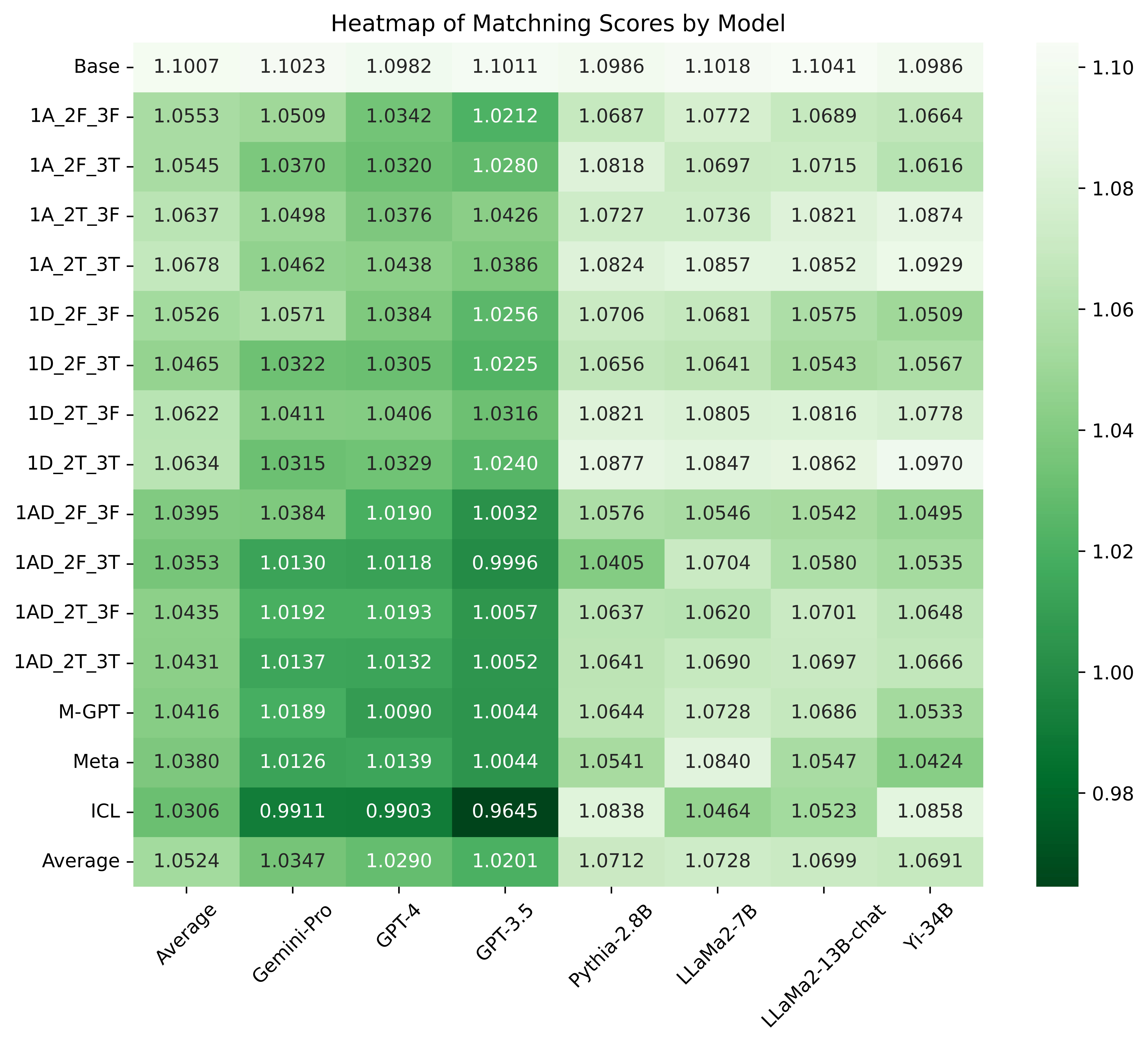} 
\caption{Matching score heatmap.}
\label{mat_detail}
\end{figure*}

\section{Details of Scores, Figure \ref{sim_detail}, \ref{mat_detail}}

\end{document}